\documentclass{article}
\usepackage{spconf,amsmath,graphicx,hyperref}
\usepackage{booktabs}
\usepackage{amssymb}
\usepackage{siunitx}
\usepackage{subcaption}
\title{Rethinking Multiplication in Neural Networks with Hadamard Transform-Assisted Operators}
\name{
Emadeldeen Hamdan$^{1}$\thanks{This work was supported by the National Science Foundation (NSF) IDEAL 2217023 and NSF POSE 2303700.},
Marouane Tliba$^{2}$,
Ulas Bagci$^{3}$,
Ahmet Enis Cetin$^{1}$
}

\address{
$^{1}$University of Illinois Chicago, USA, 
$^{2}$Universit\'e Sorbonne Paris Nord, France\\
$^{3}$Northwestern University, USA
}

\begin{document}
%
\maketitle
\begin{abstract}

Multiplication-avoiding neural operators are attractive for compute-in-memory (CIM) architectures, but directly replacing conventional convolutions often causes accuracy degradation. We investigate whether this limitation arises from multiplication avoidance itself or from the representation on which the operator acts. In this work, we introduce Hadamard Transform-assisted multiplication-avoiding (HMA) operators that combine Walsh--Hadamard transform features with multiplication-avoiding channel mixing. To identify where conventional multiplication remains necessary, we employ a gated formulation in which conventional and HMA operators compete through learned layer-wise coefficients. ResNet-18 experiments show that intermediate convolutional layers assign greater weight to HMA, whereas the earliest layers and final classifier retain greater contribution from conventional computation. Guided by this observation, we construct a compact hybrid HMA network and train it using knowledge distillation from a conventional teacher. On CIFAR-10, the resulting ResNet-18 student achieves 91.2\% accuracy compared with 90.4\% for the corresponding conventional baseline while reducing parameters by 68.25\%, conventional MACs by 85.71\%, and general multiplications by 85.54\%. Across ResNet-18, ResNet-50, and MobileNetV2 on CIFAR-10, CIFAR-100, and Tiny-ImageNet, HMA with knowledge distillation matches or improves the conventional models. These results show that Hadamard-domain feature restructuring can improve the accuracy--complexity tradeoff of multiplication-avoiding networks and provide a practical path toward efficient CIM inference.

\end{abstract}
\section{Introduction}
\label{sec:intro}

\begin{figure}[t]
    \centering
\includegraphics[width=1\columnwidth, height= 6.5cm]{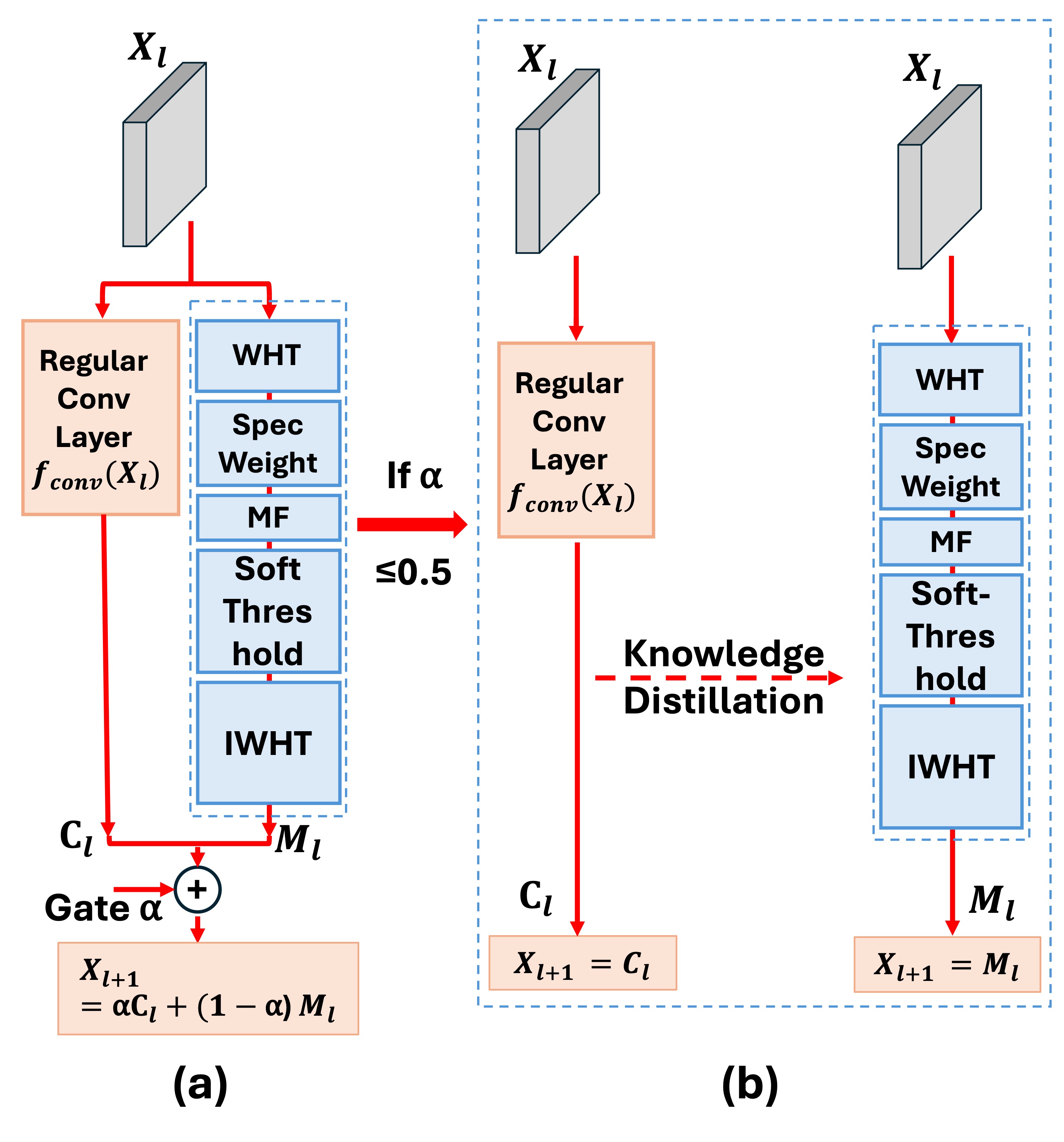}
 \caption{
Overview of the proposed HMA framework.
(a) Conventional and HMA branches are jointly trained using a layer-wise gate $\alpha_l$.
(b) Layers with $\alpha_l \leq 0.5$ are replaced by HMA, and the resulting
single-branch hybrid network is trained using knowledge distillation from the
conventional teacher for compute-in-memory inference.
}
    \label{fig:framework}
\end{figure}

The rapid growth of deep neural networks (DNNs) has intensified the demand for energy-efficient, high-throughput hardware acceleration~\cite{krizhevsky2012imagenet,he2016deep}. In convolutional neural networks (CNNs), multiply--accumulate (MAC) operations account for a substantial portion of the computational workload~\cite{lecun1998efficient,rastegari2016xnor,le2025memristor}, while multiplication is considerably more expensive than addition in conventional digital hardware~\cite{horowitz20141}. Efficient inference is further limited by the memory wall, where repeated movement of weights and activations contributes substantially to latency and energy consumption.

Compute-in-memory (CIM) architectures address this bottleneck by performing arithmetic directly within memory arrays~\cite{yu2021compute,meng2024compute,wu2025combating}. SRAM-based CIM reduces data movement and enables highly parallel computation, yet conventional neural networks remain dominated by multiplication-intensive convolutional and fully connected layers. MF-Net~\cite{nasrin2021mf} introduced multiplication-avoiding operators for SRAM-based CIM, but aggressive replacement of conventional layers can degrade predictive accuracy. This suggests that the limitation may arise not from multiplication avoidance itself, but from the representation on which such operators act.

Structured transforms provide a promising alternative. The Walsh--Hadamard Transform (WHT) is an orthogonal transform implemented using only additions and subtractions with complexity $\mathcal{O}(N\log N)$~\cite{pan2022block,pan2021fast,deveci2018energy}, and Hadamard-based layers have been explored as efficient alternatives to learned transformations~\cite{zhu2024probabilistic}. Motivated by this, we introduce a Hadamard-assisted multiplication-avoiding (HMA) operator that combines WHT-based spatial feature transformation with multiplication-avoiding channel mixing. A diagnostic gated network is used to identify where HMA can replace conventional operators. The learned gates show that intermediate convolutional layers favor HMA, whereas the earliest layers and final classifier retain stronger preference for conventional computation. In this work, we construct a compact hybrid HMA network and use knowledge distillation from a conventional teacher to overcome the optimization difficulty of aggressive operator replacement.

The main contributions are:
\begin{itemize}
\item We introduce HMA, which embeds multiplication-avoiding channel mixing
within a structured Hadamard-domain representation, substantially reducing
general multiplications and spatial convolution parameters.

\item We use a layer-wise gated routing analysis to identify where HMA can
replace conventional computation and convert the resulting diagnostic model
into a single-branch hybrid architecture.

\item Through matched routing, matched capacity, and knowledge distillation
controls, we show that the performance gain cannot be explained by routing,
model size, or distillation alone, supporting the benefit of the Hadamard
representation for multiplication-avoiding networks.
\end{itemize}

\section{Background}
\label{sec:background}

\subsection{Hadamard Transform}
\label{HT}

The Walsh--Hadamard Transform (WHT)~\cite{agaian2006hadamard,sarukhanyan2007reversible,agaian2011hadamard} is an orthogonal transform that can be computed using only additions and subtractions. For a transform size $N=2^m$, the normalized Hadamard matrix is recursively defined as
\begin{equation}
H_m =
\begin{bmatrix}
H_{m-1} & H_{m-1}\\
H_{m-1} & -H_{m-1}
\end{bmatrix},
\qquad H_0=[1].
\end{equation}
The corresponding fast WHT requires $\mathcal{O}(N\log N)$ additions/subtractions. For dyadic convolution, the WHT satisfies
$\mathrm{WHT}(a *_d x)=\mathrm{WHT}(a)\odot\mathrm{WHT}(x)$,
which motivates structured Hadamard-domain processing. Pan et al.~\cite{pan2023hybrid} exploited this property to construct HT-based convolutional layers with reduced parameter and arithmetic cost.


\subsection{Multiplication-Avoiding Operators}
Conventional convolution evaluates inner products of the form
\begin{equation}
y=\sum_{i=1}^{N} w_i x_i,
\end{equation}
which require repeated multiplication and accumulation. MF-Net~\cite{nasrin2021mf,yuan2023comprehensive} replaces the scalar product with a multiplication-avoiding operator
\begin{equation}
\label{eq:mfnet-scalar}
w\oplus x =
\operatorname{sign}(x)|w|
+
\operatorname{sign}(w)|x|,
\end{equation}
and defines the corresponding vector operation as
\begin{equation}
\label{eq:mfnet-vector}
\mathbf{w}\odot_{\mathrm{MF}}\mathbf{x}=
\sum_{i=1}^{N}
\left[
\operatorname{sign}(x_i)|w_i|
+
\operatorname{sign}(w_i)|x_i|
\right].
\end{equation}

The operator replaces general multiplications with sign, magnitude, and accumulation operations, making it attractive for multiplication-constrained and compute-in-memory implementations. MF-Net demonstrated that such operators can be mapped to SRAM-based CIM hardware; however, aggressively replacing conventional neural-network layers may reduce predictive accuracy. This motivates our investigation of whether structured Hadamard-domain representations can make multiplication-avoiding operators more effective.

\section{Proposed Method}
\label{sec:metho}

We introduce a Hadamard-assisted multiplication-avoiding (HMA) operator,
use layer-wise gating to identify where HMA can replace conventional
computation, and train the resulting hybrid network through knowledge
distillation.

\subsection{Hadamard-Assisted Multiplication-Avoiding Operator}

Let $\mathbf{X}\in\mathbb{R}^{C_{\mathrm{in}}\times H\times W}$ denote an
input feature map. For stride-preserving layers, $\mathbf{X}$ is transformed
directly; when replacing a strided convolution, non-overlapping average
pooling is first applied using the original stride. A two-dimensional WHT is
then applied independently to each channel,
\begin{equation}
\mathbf{U}=\mathcal{H}(\mathbf{X}'),
\label{eq:wht_forward}
\end{equation}
where $\mathbf{X}'$ denotes the original or downsampled feature map.
Spatial dimensions are padded to the next power of two when necessary and
cropped after the inverse transform.

The transformed representation is modulated by a learnable spectral weighting
matrix $\mathbf{V}$,
\begin{equation}
\tilde{\mathbf{U}}=\mathbf{V}\odot\mathbf{U},
\label{eq:spectral_gate}
\end{equation}
where $\mathbf{V}\in\mathbb{R}^{H'\times W'}$ is shared across channels.
Channel mixing is performed using the multiplication-avoiding operator in
Eq.~\eqref{eq:mfnet-scalar}. For output channel $j$,
\begin{equation}
\begin{aligned}
Z_j(h,w)=\sum_{i=1}^{C_{\mathrm{in}}}\Big[
&\operatorname{sign}\!\left(\tilde U_i(h,w)\right)|W_{j,i}|\\
&+\operatorname{sign}(W_{j,i})|\tilde U_i(h,w)|
\Big].
\end{aligned}
\label{eq:HMA_mixing}
\end{equation}
The resulting channel mixer contains
$C_{\mathrm{in}}C_{\mathrm{out}}$ learned weights rather than
$K^2C_{\mathrm{in}}C_{\mathrm{out}}$ for a conventional $K\times K$
convolution.

A learnable soft-thresholding operation is then applied,
\begin{equation}
\mathcal{S}(z;\tau)=
\operatorname{sign}(z)\max(|z|-\tau,0),
\label{eq:soft_threshold}
\end{equation}
followed by the inverse WHT,
\begin{equation}
\mathbf{Y}_{\mathrm{HMA}}
=
\mathcal{H}^{-1}\!
\left(\mathcal{S}(\mathbf{Z};\boldsymbol{\tau})\right)
+\mathbf{b}.
\label{eq:HMA_final}
\end{equation}
Because the learned spectral weighting in Eq.~\eqref{eq:spectral_gate}
retains conventional multiplications, we describe HMA as
\emph{multiplication-avoiding} rather than completely multiplication-free.

\subsection{Gated Operator Selection}

To determine where conventional computation remains necessary, each eligible
layer is represented by parallel conventional and HMA branches,
\begin{equation}
\mathbf{X}_{l+1}
=
\alpha_l f_l^{\mathrm{conv}}(\mathbf{X}_l)
+
(1-\alpha_l)f_l^{\mathrm{HMA}}(\mathbf{X}_l),
\label{eq:gated_layer}
\end{equation}
where
\begin{equation}
\alpha_l=\sigma(a_l), \qquad a_l=0\;\text{at initialization}.
\label{eq:alpha}
\end{equation}
Thus, $\alpha_l\rightarrow1$ indicates preference for conventional
computation, while $\alpha_l\rightarrow0$ favors HMA. The gates are used
only for operator selection and are removed from the final model.

As illustrated in Fig.~\ref{fig:framework}, the gated network is converted
into a single-branch hybrid architecture by selecting HMA when
$\alpha_l\leq0.5$ and retaining the conventional operator otherwise.

\subsection{Hybrid Training with Knowledge Distillation}

The selected hybrid architecture is reinitialized and trained independently.
Since direct optimization produces a substantial accuracy gap, we use the
original conventional network as a teacher. Let $\mathbf{z}_T$ and
$\mathbf{z}_S$ denote teacher and student logits. The student is optimized as
\begin{equation}
\begin{aligned}
\mathcal{L}
=&\;
(1-\lambda)\mathcal{L}_{\mathrm{CE}}
(\mathbf{y},\mathbf{z}_S)\\
&+\lambda T^2D_{\mathrm{KL}}
\left(
\sigma(\mathbf{z}_T/T)
\Vert
\sigma(\mathbf{z}_S/T)
\right).
\end{aligned}
\label{eq:kd}
\end{equation}
Here, $T$ is the distillation temperature and $\lambda$ controls the
teacher contribution. The teacher is used only during training; inference
uses only the selected hybrid HMA network.

\section{Experimental Results}
\label{sec:results}

\subsection{Experimental Setup}

We evaluate HMA using ResNet-18, ResNet-50~\cite{he2016deep}, and MobileNetV2~\cite{howard2017mobilenets} on
CIFAR-10, CIFAR-100~\cite{krizhevsky2009learning}, and Tiny ImageNet~\cite{deng2009imagenet}. All models are trained for
100 epochs using SGD with momentum $0.9$, an initial learning rate of
$0.01$, weight decay $5\times10^{-5}$, and batch size $128$. Training uses zero-padded random
cropping and horizontal flipping with probability $0.5$, followed by
normalization using training-set statistics. Model selection uses
class-stratified $45{,}000/5{,}000$ training/validation splits for
CIFAR-10 and CIFAR-100 and a $90{,}000/10{,}000$ split for
Tiny-ImageNet. The official test sets are used only for final evaluation.
Principal results are reported as mean $\pm$ standard deviation over
three random seeds.

For knowledge distillation, we use temperature $T=4$ and distillation
weight $\lambda=0.5$, with objective
\begin{equation}
\mathcal{L}
=
(1-\lambda)\mathcal{L}_{\mathrm{CE}}
+
\lambda T^2\mathcal{L}_{\mathrm{KL}}.
\end{equation}
The conventional teacher checkpoint is selected according to the highest
validation accuracy on the held-out selection split. Gate logits are
initialized as $a_l=0$, giving $\alpha_l=\sigma(a_l)=0.5$. Routing is
determined from the trained gated model using $\alpha_l<0.50$ for HMA
and conventional computation otherwise. The HMA soft-threshold parameters
are initialized independently from $\mathcal{U}(0,0.1)$, with effective
thresholds constrained to be nonnegative using
$\tau_l^{+}=\operatorname{ReLU}(\tau_l)$. Threshold parameters are
excluded from weight decay.

ResNet-18 is used for detailed ablation. The resulting routing pattern is
transferred to ResNet-50 and MobileNetV2; for MobileNetV2, depthwise
convolutions remain conventional while eligible channel-mixing layers are
replaced by HMA.

Operation counts include the arithmetic of the WHT/IWHT, learned spectral
weighting, average-pooling adapters for strided layers,
multiplication-avoiding channel mixing, and all retained conventional
operators. Memory movement, latency, energy, and hardware-specific
peripheral overhead are not modeled. The experiments are designed as controlled operator comparisons rather than
state-of-the-art accuracy benchmarks. Within each backbone and dataset,
models use the same preprocessing, optimization schedule, and training
budget unless otherwise stated for knowledge distillation.

\subsection{Main Results}

Table~\ref{tab:resnet18_main} compares the conventional ResNet-18 with the
final distilled HMA hybrid on CIFAR-10. HMA improves accuracy from 90.4\% to
91.2\% while reducing parameters from 11.174M to 3.548M (68.25\%) and general
multiplications from 556.04M to 80.39M (85.54\%). Conventional MACs are
similarly reduced by 85.71\%.

Operation counts include WHT/IWHT arithmetic, spectral weighting,
average-pooling adapters, MF channel mixing, and retained conventional
operators; they do not model memory traffic, latency, or CIM energy. Therefore, HMA reduces both learned parameters and general multiplications
without sacrificing accuracy. The additional sign and magnitude operations
are reported explicitly rather than absorbed into a conventional FLOP count. 

\begin{table}[ht]
\centering
\caption{ResNet-18 accuracy and algorithmic arithmetic complexity on
CIFAR-10. Operation counts are reported per $32\times32$ input image and
exclude hardware-specific memory, latency, and energy costs.}
\label{tab:resnet18_main}

\begin{tabular}{lcc}
\hline
\textbf{Metric} & \textbf{ResNet-18} & \textbf{Hybrid HMA} \\
\hline
Accuracy (\%) & 90.4 & \textbf{91.2} \\
Parameters & 11.174M & \textbf{3.548M} \\
Parameter storage & 42.63 MiB & \textbf{13.53 MiB} \\
Conventional MACs & 555.42M & \textbf{79.37M} \\
General multiplications & 556.04M & \textbf{80.39M} \\
Additions/subtractions & 555.68M & \textbf{200.45M} \\
Sign evaluations & 0 & 113.69M \\
Absolute-value operations & 0 & 113.69M \\
Sign applications & 0 & 113.69M \\
\hline
\end{tabular}
\end{table}

Table~\ref{tab:backbones} evaluates whether this behavior generalizes beyond
ResNet-18. Across ResNet-18, ResNet-50, and MobileNetV2, HMA+KD matches or
improves the corresponding conventional model on all three datasets. The
largest gains are observed on CIFAR-10, with improvements of 0.8, 0.7, and
0.9 percentage points for ResNet-18, ResNet-50, and MobileNetV2,
respectively.

\begin{table}[t]
\centering
\caption{Generalization of the proposed HMA approach across network architectures and datasets. C10 and C100 denote CIFAR-10 and CIFAR-100, respectively, while Tiny-IN denotes Tiny-ImageNet. Results are reported as mean $\pm$ standard deviation over three seeds. }
\label{tab:backbones}
\resizebox{\columnwidth}{!}{
\begin{tabular}{llccc}
\hline
\textbf{Backbone} & \textbf{Method}
& \textbf{C10} & \textbf{C100} & \textbf{Tiny-IN} \\
\hline
ResNet-18~\cite{he2016deep}
& Conventional & 90.4$\pm$0.1 & 65.0$\pm$0.2 & 55.3$\pm$0.2 \\
& \textbf{HMA + KD} & \textbf{91.2$\pm$0.2} & \textbf{65.9$\pm$0.2} & \textbf{56.1$\pm$0.1} \\
\hline
ResNet-50~\cite{he2016deep}
& Conventional & 93.3$\pm$0.2& 77.0$\pm$0.3 & 65.1$\pm$0.4 \\
& \textbf{HMA + KD} & \textbf{94.0$\pm$0.3} & \textbf{77.6$\pm$0.4} & \textbf{65.8$\pm$0.4} \\
\hline
MobileNetV2~\cite{howard2017mobilenets}
& Conventional & 88.8$\pm$0.2 & 61.0$\pm$0.1 & 56.8$\pm$0.3 \\
& \textbf{HMA + KD }& \textbf{89.7$\pm$0.2} & \textbf{61.4$\pm$0.1} & \textbf{57.5$\pm$0.2} \\
\hline
\end{tabular}}
\end{table}

On CIFAR-100, HMA improves ResNet-18, ResNet-50, and MobileNetV2 by 0.9, 0.6, and 0.4 points,
respectively. On Tiny-ImageNet, the
improvements are 0.8, 0.7, and 0.7 points across the three backbones. These
results indicate that the HMA strategy transfers across standard, deeper, and
efficiency-oriented CNN architectures.

\subsection{Ablation Studies}


\noindent\textbf{Hadamard Representation.} To isolate the contributions of the Hadamard representation and
multiplication-avoiding arithmetic, we compare conventional convolution, direct MF, HT with conventional channel mixing, and HMA. On CIFAR-10, the conventional ResNet-18 achieves 90.4\% accuracy. Direct MF achieves 62.9\%, while HT with conventional channel mixing reaches 88.3\%. Under identical hybrid routing and knowledge distillation, MF reaches 78.1\%, whereas HMA achieves 91.2\%, which uses Hadamard-domain restructuring substantially in multiplication-avoiding channel mixing.


\noindent\textbf{Selection and Distillation.}
The learned gates exhibit a clear depth-dependent pattern, with mean
$\bar{\alpha}=0.40$. The first two layers reach $\alpha=0.59$ and
$\alpha=0.60$, and the final fully connected layer reaches $\alpha=0.97$;
these layers are retained as conventional operators, while most intermediate
layers are replaced by HMA. Demonstrated in
Table~\ref{tab:selection_ablation}, directly training the resulting hard-routed HMA
model achieves 77.6\%, but knowledge distillation increases accuracy to
91.2\%. Under the same routing, MF achieves 63.7\% without KD and 78.1\%
with KD. A capacity-matched conventional CNN with approximately 3.55M
parameters achieve 67.0\% without KD and 73.0\% with KD. These controls
indicate that the HMA gain cannot be explained by routing, model capacity,
or distillation alone.

\begin{table}[t]
\centering
\caption{Effect of operator selection and knowledge distillation on ResNet-18/CIFAR-10.}
\label{tab:selection_ablation}
\begin{tabular}{lccc}
\hline
\textbf{Model} & \textbf{KD} & \textbf{Routing} & \textbf{ACC (\%)} \\
\hline
Conventional ResNet-18 & -- & -- & 90.4 \\
MF & -- & Full & 62.9 \\
MF & -- & Same & 63.7 \\
MF & \checkmark & Same & 78.1 \\
Gated Conv/HMA & -- &  Soft &90.3 \\
Hard hybrid HMA & -- &  Same &77.6 \\
\textbf{Hard hybrid HMA} & \checkmark &  Same & \textbf{91.2} \\
\hline
\end{tabular}
\end{table}

\section{Conclusion}

We introduced a Hadamard-assisted multiplication-avoiding (HMA) framework
for reducing general multiplications in CNNs while preserving predictive
performance. Layer-wise gating showed that intermediate convolutions assign
greater weight to HMA, whereas early layers and the classifier retain greater
contribution from conventional computation. Across ResNet-18, ResNet-50, and
MobileNetV2, HMA with knowledge distillation matched or improved the
corresponding conventionally trained models on CIFAR-10, CIFAR-100, and
Tiny-ImageNet. On ResNet-18, the distilled HMA student achieved 91.2\%
accuracy compared with 90.4\% for the conventional teacher while reducing
parameters by 68.25\% and general multiplications by 85.54\%. These results
support HMA as a promising multiplication-reduction strategy for
compute-in-memory-oriented inference.

\bibliographystyle{IEEEbib}
\bibliography{strings,refs}

\end{document}